\documentclass[10pt,twocolumn,letterpaper]{article}

\usepackage{iccv}
\usepackage{times}
\usepackage{epsfig}
\usepackage{graphicx}
\usepackage{amsmath}
\usepackage{amssymb}
\usepackage{booktabs}
\usepackage{multirow}
\usepackage{caption} 
\usepackage{graphicx}
\usepackage{subcaption}
\usepackage{authblk}
\usepackage{lipsum}

\usepackage[breaklinks=true,bookmarks=false]{hyperref}

\iccvfinalcopy 

\def\iccvPaperID{****} 

\begin{document}

\title{Evaluating and Predicting Distorted Human Body Parts for Generated Images}

\author[1,2]{Lu Ma\textsuperscript{*}\textsuperscript{†}}
\author[2]{Kaibo Cao\textsuperscript{*}}
\author[1]{Hao Liang}
\author[2]{Jiaxin Lin}
\author[2]{Zhuang Li}
\author[2]{Yuhong Liu}
\author[2]{Jihong Zhang}
\author[1]{Wentao~Zhang\textsuperscript{‡} }
\author[1]{Bin Cui\textsuperscript{‡}}


\affil[1]{Peking University}
\affil[2]{Tencent Inc.}



\maketitle

\let\thefootnote\relax\footnotetext{* Equal contribution}
\let\thefootnote\relax\footnotetext{† The work was done during an internship at Tencent \\ Email: \textsl{maluqaq@163.com}}
\let\thefootnote\relax\footnotetext{‡ Corresponding author \\ Email: \{\textsl{wentao.zhang},\textsl{bin.cui}\}\textsl{@pku.edu.cn}}

\begin{abstract}
Recent advancements in text-to-image (T2I) models enable high-quality image synthesis, yet generating anatomically accurate human figures remains challenging. AI-generated images frequently exhibit distortions such as proliferated limbs, missing fingers, deformed extremities, or fused body parts. Existing evaluation metrics like Inception Score (IS) and Fréchet Inception Distance (FID) lack the granularity to detect these distortions, while human preference-based metrics focus on abstract quality assessments rather than anatomical fidelity. To address this gap, we establish the first standards for identifying human body distortions in AI-generated images and introduce Distortion-5K, a comprehensive dataset comprising 4,700 annotated images of normal and malformed human figures across diverse styles and distortion types. Based on this dataset, we propose ViT-HD, a Vision Transformer-based model tailored for detecting human body distortions in AI-generated images, which outperforms state-of-the-art segmentation models and visual language models, achieving an F1 score of 0.899 and IoU of 0.831 on distortion localization. Additionally, we construct the Human Distortion Benchmark with 500 human-centric prompts to evaluate four popular T2I models using trained ViT-HD, revealing that nearly 50\% of generated images contain distortions. This work pioneers a systematic approach to evaluating anatomical accuracy in AI-generated humans, offering tools to advance the fidelity of T2I models and their real-world applicability. The Distortion-5K dataset, trained ViT-HD will soon be released in our GitHub repository: \href{https://github.com/TheRoadQaQ/Predicting-Distortion}{https://github.com/TheRoadQaQ/Predicting-Distortion}.
\end{abstract}

\begin{figure*}[!ht]
\centering
\includegraphics[width=0.98\textwidth]{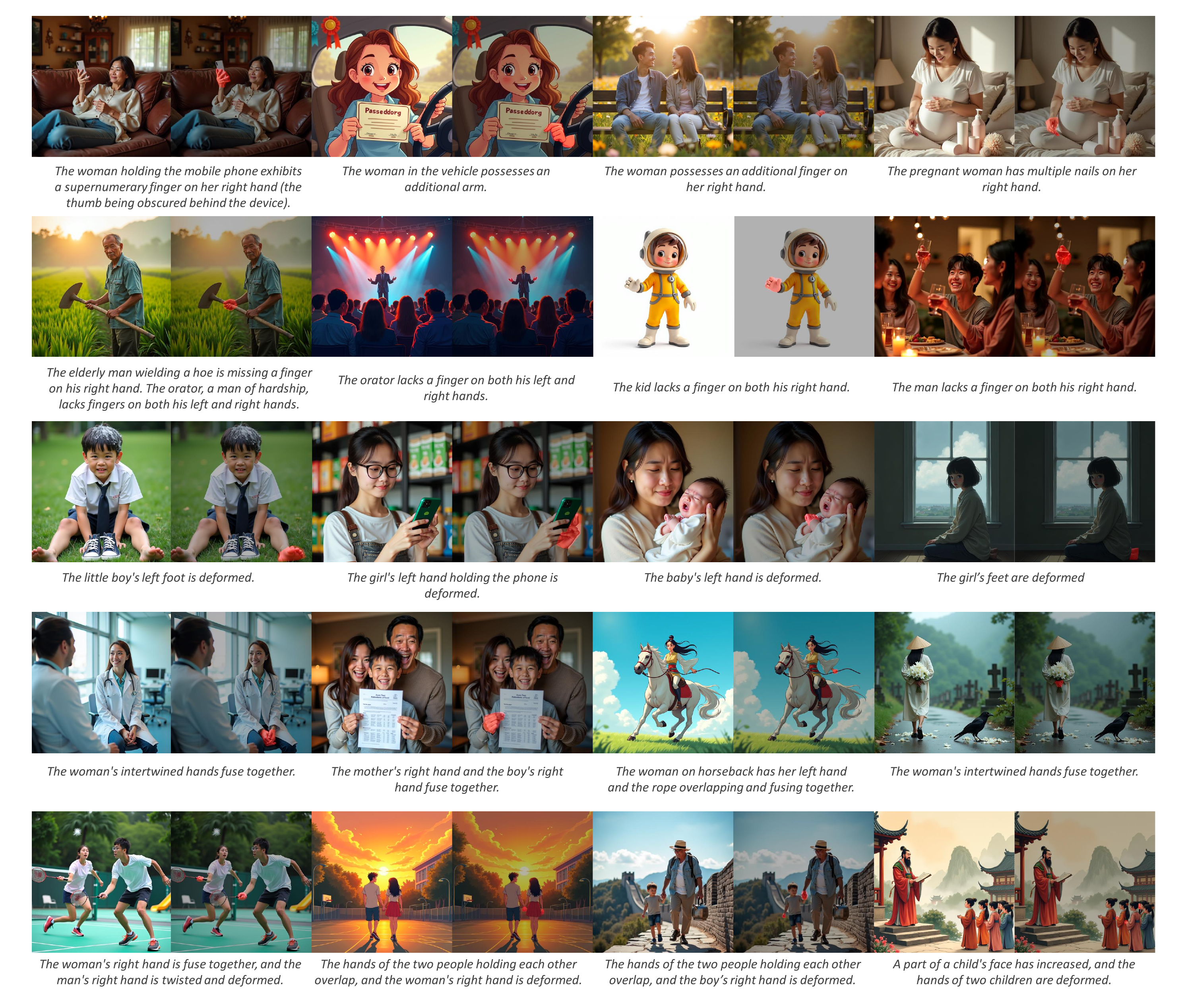}
\caption{\textbf{Examples from our \emph{Distortion-5K}}. AI-generated human images frequently exhibit various distortions, including proliferation (first row), absence (second row), deformation (third row), fusion (fourth row), and the occurrence of multiple distortions within a single image (fifth row). We annotate the distorted body parts in these images, where the left image in each pair represents the original, and the right image features red masks indicating the distorted regions.
}
\label{fig:distortion_examples}
\end{figure*}

\section{Introduction}
In recent years, text-to-image (T2I) models achieve remarkable advancements, enabling the synthesis of high-quality images directly from textual descriptions~\cite{yang2023diffusion,dhariwal2021diffusion,chen2023pixart,rombach2022high}. These models demonstrate vast potential across a wide range of applications, including digital art creation~\cite{wang2024diffusion}, virtual reality~\cite{herz2019understanding}, and automated content generation~\cite{chen2024overview}.

Despite these significant strides, AI-generated images still face notable challenges, particularly in rendering human figures with anatomical accuracy. The synthesis of human bodies remains a complex task due to the intricate anatomical structures and subtle nuances involved, especially in the depiction of hands and feet~\cite{gandikota2025concept, lu2024handrefiner}. As illustrated in Figure \ref{fig:distortion_examples}, AI-generated human images often exhibit distortions such as the proliferation of hands, the absence of fingers, the deformation of feet, and the fusion of body parts. We define such instances as \textbf{distorted}, and our experimental results indicate that nearly 50\% of human-related generated images suffer from these distortions, highlighting a critical limitation in current T2I models.

Existing evaluation metrics for generated images are inadequate for detecting these specific issues related to human body distortions. Metrics such as the Inception Score (IS)~\cite{barratt2018note} and the Fréchet Inception Distance (FID)~\cite{heusel2017gans} often fail to align with human judgment~\cite{ding2022cogview2} and lack the granularity needed to identify distorted human images. While metrics derived from human preferences provide summarized numeric scores of image quality~\cite{wu2023human,wu2023humanv2}, they are limited to broad assessments and fail to address finer details or provide insights into specific areas where a model underperforms, particularly in the integrity of human anatomical structures in generated images.

To address this gap, we establish standards for identifying human body distortions and introduce a comprehensive dataset of normal and distorted AI-generated human images. Additionally, we present a distortion detection model, specifically designed to evaluate and predict these distortions. Our contributions can be summarized as follows:

\begin{itemize}
\item We establish the first comprehensive set of standards for identifying human body distortions in AI-generated images and introduce the \emph{Distortion-5K} dataset, which contains approximately 5,000 images of normal and various types of distorted human figures, including detailed annotations of distorted regions.

\item We propose a \textbf{Vi}sion \textbf{T}ransformer model for \textbf{H}uman \textbf{D}istortion in AI-generated images, termed ViT-HD, which is capable of evaluating and identifying various types of distortions across different body parts in a wide range of artistic styles.

\item We introduce the \emph{Human Distortion Benchmark} and conduct an extensive evaluation of popular T2I models using our ViT-HD model. This evaluation involves analyzing the frequency of human body distortions in generated images, providing valuable insights into the performance and limitations of these models in rendering anatomically accurate human figures.
\end{itemize}

To the best of our knowledge, this work represents the first dedicated effort to develop both a specialized model and a dataset focused on evaluating distortions in AI-generated human figures. We believe that our contributions significantly advance the field of AI image generation, offering a pathway towards more precise and anatomically accurate synthesis of human bodies. Our work not only addresses a critical gap in current evaluation methods but also provides a valuable toolset for researchers and practitioners aiming to improve the fidelity of T2I models in human figure generation.
\section{Related Works}
\subsection{Text-to-image Generation}
Text-to-image generation sees significant advancements through various model architectures. Initially, Generative Adversarial Networks (GANs)~\cite{goodfellow2014generative} use a generator and discriminator to produce and evaluate images. However, GANs struggle with generating consistently diverse and high-quality images. This leads to the exploration of Variational Auto-Encoders (VAEs)~\cite{van2017neural, kingma2013auto}, which optimize the evidence lower bound to improve image likelihood.

The introduction of diffusion models~\cite{ho2020denoising} marks a significant shift, setting a new state-of-the-art by iteratively refining images from random noise, thus capturing greater image diversity and quality. Latent Diffusion Models (LDMs)~\cite{rombach2022high} further improve efficiency by performing the diffusion process in a compact latent space. Additionally, pixel-based models like DALL-E~\cite{ramesh2021zero} and Imagen~\cite{saharia2022photorealistic} achieve superior text-to-image alignment and resolution.

\subsection{Text-to-image Evaluation}
Inception Score (IS)~\cite{barratt2018note} and Fréchet Inception Distance (FID)~\cite{heusel2017gans} are commonly used to evaluate text-to-image models but fall short in assessing single images and aligning with human preferences. Recent studies focus on collecting and learning human preferences to fine-tune visual language models (VLMs).

Wu et al.~\cite{wu2023human,wu2023humanv2} create a dataset of 98,807 images based on user prompts and train a model to predict the Human Preference Score (HPS). Kirstain et al.~\cite{kirstain2023pick} develop a web app to gather prompts and user preferences, resulting in the Pick-a-Pic dataset, which focuses on overall preferences but lacks detailed annotations. Xu et al.~\cite{xu2024imagereward} collect a dataset by having users rank and rate images on quality, alignment, and fidelity, and train the ImageReward model for human preference learning. AGIQA-20k~\cite{li2024aigiqa} compiles 20,000 AI-generated images with 420,000 subjective scores on perceptual quality and text-to-image alignment. Liang et al.~\cite{liang2024rich} enrich feedback by annotating implausible/misaligned image regions and misrepresented/missing words in text prompts, using data from 18,000 images to train a feedback prediction model. Zhang et al.~\cite{zhang2024learning} introduce the Multi-dimensional Preference Score using their Multi-dimensional Human Preference Dataset to evaluate text-to-image models across multiple dimensions.

Despite these valuable contributions, most existing works focus on learning from human perceptual preferences for all types of AI-generated images through abstract dimensions (e.g., text-to-image alignment, aesthetics, and fidelity). However, there is a noticeable gap in research specifically aimed at evaluating and detecting distortions in AI-generated human figures, despite its critical importance. One work relevant to our work is~\cite{chen2023exploring}, which attempts to benchmark and assess the visual naturalness of AI-generated images that implicitly encompass the concept of human distortion. Nevertheless, their research primarily addresses the broader concept of naturalness, whereas our study specifically targets distorted human body regions. To the best of our knowledge, this is the first work focused on evaluating and detecting human body distortions in AI-generated images.

\begin{figure*}[htbp]
\centering
\begin{subfigure}[b]{0.29\textwidth}
\centering
\includegraphics[width=\linewidth]{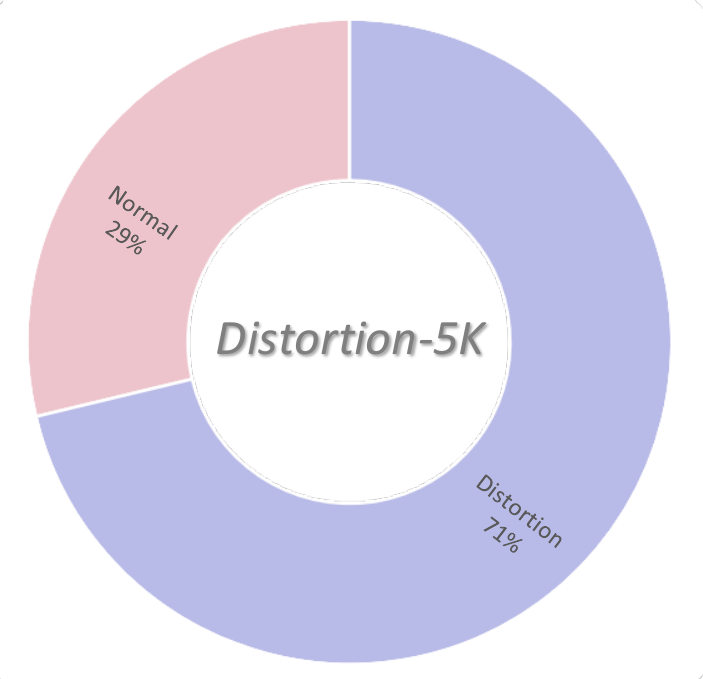}
\caption{The rate of images containing distorted regions.}
\label{fig:distortion_distribution}
\end{subfigure}
\begin{subfigure}[b]{0.29\textwidth}
\centering
\includegraphics[width=\linewidth]{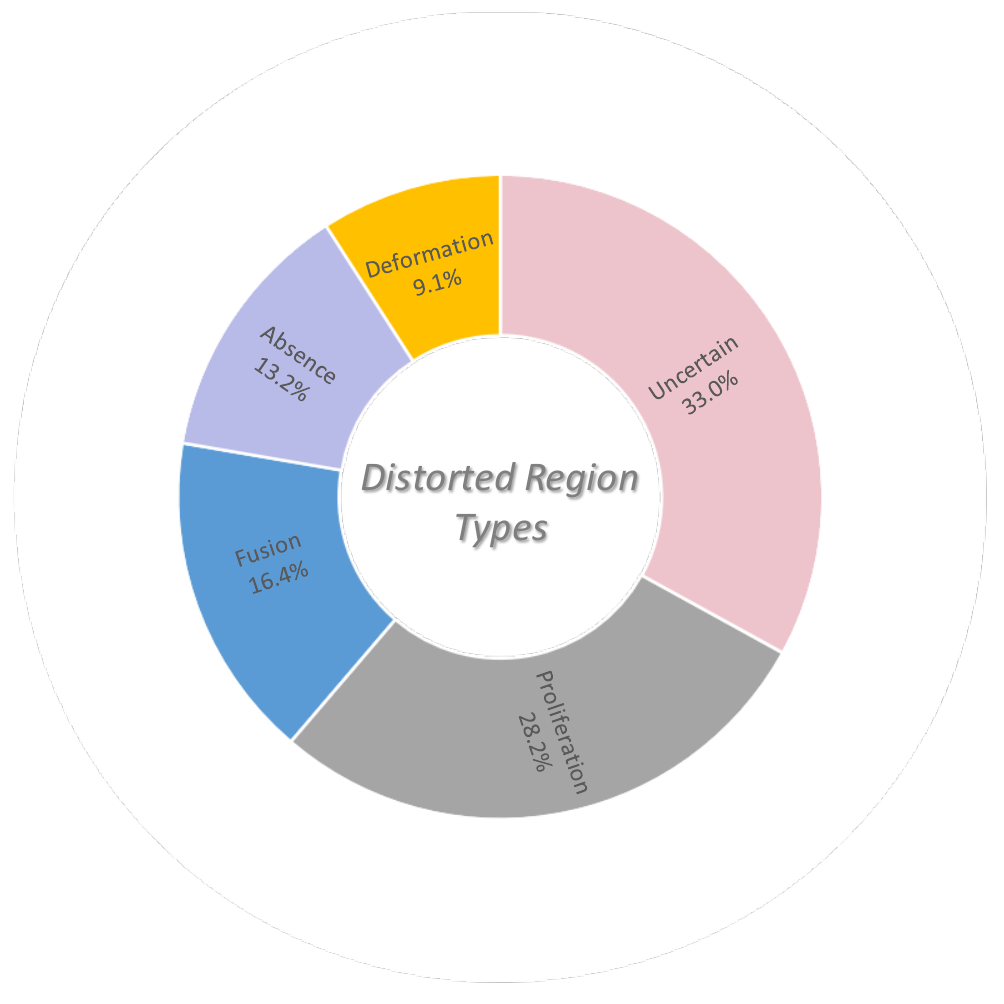}
\caption{The distribution of types of distorted regions.}
\label{fig:distortion_type_distribution}
\end{subfigure}
\begin{subfigure}[b]{0.39\textwidth}
\centering
\includegraphics[width=\linewidth]{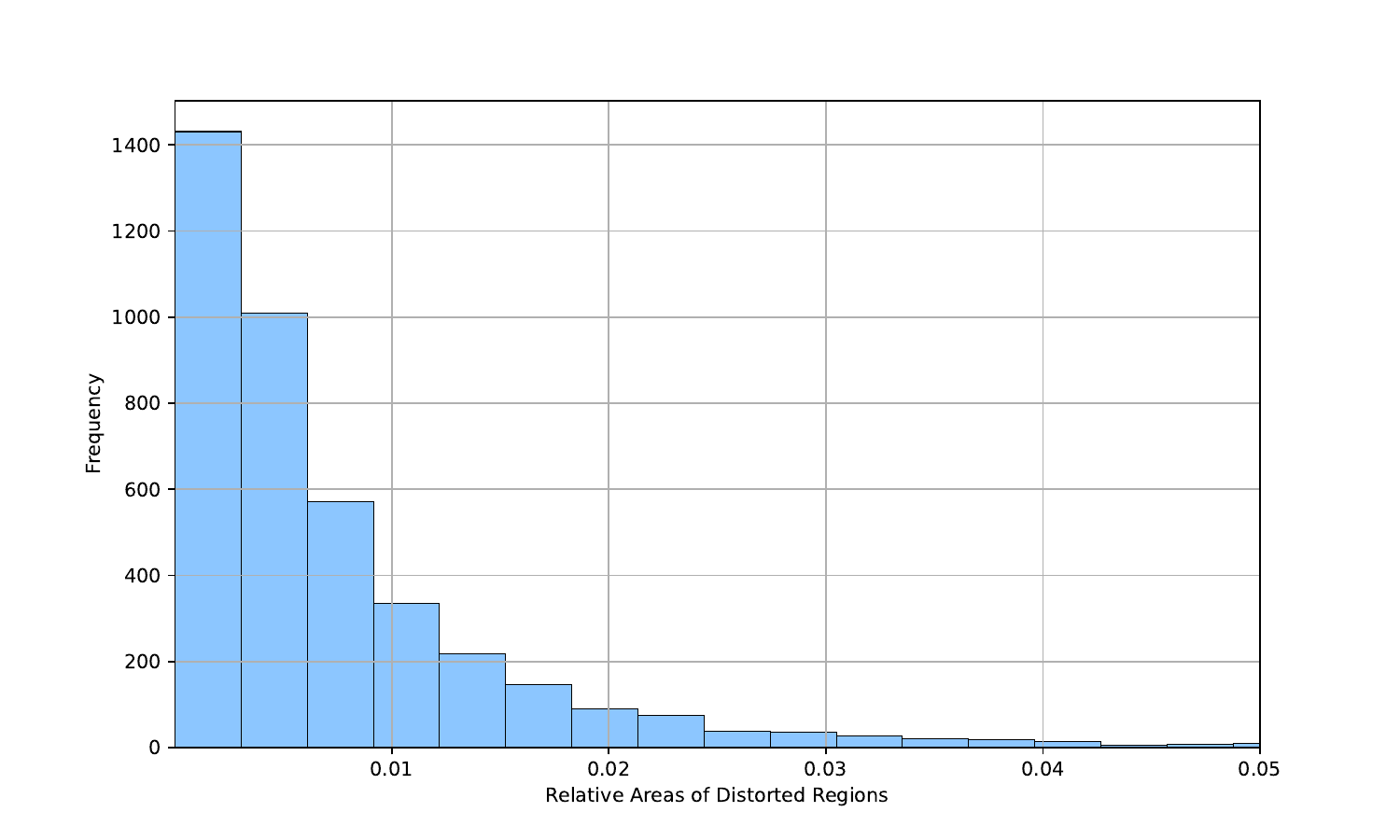}
\caption{The frequency of the relative areas of distorted regions in \emph{Distortion-5K}}
\label{fig:dataset_frequency}
\end{subfigure}
\caption{\textbf{Analysis of our \emph{Distortion-5K} dataset.} Left: The rate of distorted human images. Mid: The distribution of distortion types. Right: The frequency of the relative distorted areas.}
\label{fig:Distortion-5K}
\end{figure*}

\subsection{Plausible Human Body Generation}
Several studies investigate the challenges posed by distortion in AI-generated human figures, particularly hands~\cite{gandikota2025concept, lu2024handrefiner}. Concept Sliders~\cite{gandikota2025concept} uses parameter-efficient training to improve hand anatomy by fine-tuning diffusion models with a curated dataset. Ye et al.~\cite{ye2023affordance} generate hand-object interactions in images by guiding the process with palm and forearm areas, while HanDiffuser~\cite{ye2023affordance} incorporates 3D hand parameters into text embeddings for realistic hand depictions. Diffusion-HPC~\cite{weng2023diffusion} enhances malformed human depictions using a conditional diffusion model and depth images from reconstructed human meshes. HandRefiner~\cite{lu2024handrefiner} and RHanDS~\cite{wang2024rhands} offer post-processing techniques specifically for correcting hand deformities.

Although these methods generate nearly plausible human images or correct distortions in human hands, they do not focus on the localization and evaluation of various human body distortions. Moreover, they primarily address cases with clearly visible, realistic-style hand distortions where the hands occupy a large area of the image, which are not broadly applicable to the diverse requirements of real-world scenarios. Our method not only enables the detection of distortions across various human body parts (such as hands and feet) but also accommodates complex scenarios, including situations where hands are obscured or multiple hands appear in a single image. Additionally, it is versatile enough to handle a diverse range of styles, from photorealistic to animated representations.

\section{Collecting Datasets}
This section outlines the procedure for collecting, annotating, and constructing the \emph{Distortion-5K} dataset, which comprises nearly 5,000 (specifically, 4,700) AI-generated images of humans exhibiting various distortions, diverse styles, and varying numbers of individuals in each image.

\subsection{Data Collection Process}
To accurately identify and label distorted regions within human images, we implement a meticulous annotation process. Annotators are instructed to use polygonal shapes to outline distorted parts of the human body in each image. Additionally, for each distorted region, annotators select the type of distortion from a predefined set of categories, including proliferation, absence, deformation, and fusion. This dual approach of spatial delineation and distortion categorization is designed to assist annotators in achieving precise and consistent annotations, thereby enhancing the overall quality of the dataset.

\begin{table*}[h]
\centering
\captionsetup{font=normalsize} 
\caption{\textbf{Comparison of different distortion detection models on the \emph{Distortion-5K} test set.} Qwen2-VL(sft) represents the Qwen2-VL fine-tuned on the \emph{Distortion-5K} train set. ViT-HD(PLT) represents the model trained with pixel-level mask training. ViT-HD(TST) represents the model trained with two-stage training, first with patch-level mask training and then with pixel-level mask training.}
\begin{tabular}{lcccccccc}
\toprule
\multirow{2}{*}{\textbf{Model}} & \multicolumn{3}{c}{\textbf{Pixel-Level metrics}} & \multicolumn{2}{c}{\textbf{Area-Level metrics}} & \multicolumn{3}{c}{\textbf{Image-Level metrics}} \\
\cmidrule(lr){2-4} \cmidrule(lr){5-6} \cmidrule(lr){7-9}
 & Precision & Recall & F1 & IoU & Dice & Precision & Recall & F1 \\
\midrule
U-Net & 0.502 & 0.505 & 0.501 & 0.492 & 0.501 & 0.69 & \textbf{1.0} & 0.820\\
Deeplabv3 & 0.726 & 0.704 & 0.715 & 0.635 & 0.715 & 0.715 & 0.919 & 0.804\\
CLIP & 0.704 & 0.648 & 0.671 & 0.601 & 0.671 & 0.705 & 0.930 & 0.812\\
DINO-v2 & 0.701 & 696 & 0.698 & 0.621 & 0.698 & 0.713 & 0.761 & 0.736 \\
GPT-4o & - & - & - & - & - &  0.481 & 0.405 & 0.246\\
Qwen2-VL & - & - & - & - & - & 0.708 & \textbf{1.0} & 0.829 \\
Qwen2-VL(sft) & - & - & - & 0.307 & 0.347 & \textbf{0.849} & 0.640 & 0.730\\
ViT-HD(PLT)& 0.686 & 0.758 & 0.716 & 0.635 & 0.716 & 0.708 & 0.954 & 0.813 \\
ViT-HD(TST)& \textbf{0.905} & \textbf{0.893} & \textbf{0.899} & \textbf{0.831} & \textbf{0.899} & 0.830 & 0.945 & \textbf{0.884} \\
\bottomrule
\end{tabular}
\label{tab:ViT-HD}
\end{table*}

\subsection{Distorted Area Consolidation}
To ensure reliable and accurate annotation of distorted regions, we use a consensus-based consolidation approach. Each image is independently annotated by three different annotators, and we consolidate these annotations by selecting pixels marked as distorted by at least two annotators. This method leverages the combined expertise of multiple annotators to produce more precise annotations. The rationale for this strategy is twofold:

\begin{itemize}
\item \textbf{Validation of Annotation Quality:} We randomly sample 100 images from the dataset and have an additional expert review the annotations. The expert finds that over 90\% of these samples are correctly annotated, and nearly 5\% are ambiguous cases (which can be annotated or not), demonstrating that selecting pixels marked by at least two annotators effectively captures the true distorted regions.
\item \textbf{Enhancement of Annotation Precision:} Individual polygon annotations may lack precision due to annotator variability or the complexity of outlining distorted regions. By considering the intersection of annotations from multiple annotators, we reduce individual errors and achieve a more accurate region of distorted areas.
\end{itemize}

Consequently, each image in the \emph{Distortion-5K} dataset is labeled with a binary mask indicating the pixels belonging to distorted regions. This labeling method ensures that the dataset provides precise annotations suitable for training and evaluating models designed to detect distorted human body regions.

\subsection{Distortion-5K: A Dataset of Distorted Human Images}
The Distortion-5K dataset is compiled by collecting images from various publicly available sources, ensuring diversity in terms of environments, human poses, and types of distortions. We perform several preprocessing steps to prepare the dataset, such as filtering images that do not contain human figures or contain objects resembling humans (e.g., mannequins, statues) using trained models.

We randomly split the approximately 5K samples into three subsets: a training set with 4,000 samples, a validation set with 300 samples, and a test set with 400 samples. Additionally, the test set is double-checked by expert annotators to ensure the highest possible annotation quality. This set is reserved for the final evaluation of model performance.

To understand the characteristics of the \emph{Distortion-5K} dataset, we conducted a quantitative analysis using three key metrics derived from the mask labels. As shown in Figure \ref{fig:distortion_distribution}, we calculated the ratio of images with distorted regions (\emph{positive samples}) to those without (\emph{negative samples}), finding that negative samples make up 28\% of the dataset. We also analyzed the distribution of each distortion type in Figure \ref{fig:distortion_type_distribution}. Pixels marked by at least two annotators effectively capture true distorted regions. If all annotators agree on the distortion type, it is classified accordingly; otherwise, it is labeled as \textbf{uncertain}. Despite cross-annotation ensuring quality, 33\% of distorted regions remain uncategorized. This is due to the inherent complexity of the distortions, which may reasonably belong to multiple categories. Our main goal is to distinguish between normal and distorted regions, so specific distortion types are not considered in subsequent tasks. Additionally, we analyzed the relative areas (as a percentage of total image area) of the annotated distorted regions, with the detailed distribution shown in Figure \ref{fig:dataset_frequency}. The areas of distorted regions follow a long-tail distribution, with most distortions concentrated in smaller areas, demanding high precision from the model to capture fine details.
\section{Predicting Distorted Human Body Parts}
\subsection{Model Architecture}
We adopt the Qwen2-VL-Instruct~\cite{qwen2-vl} model, renowned for its superior performance and high-resolution capabilities. Specifically, we extract only the vision encoder component of Qwen2-VL and enhance it by integrating a multi-layer perceptron (MLP) head to predict distortion masks.

The primary motivations for selecting Qwen2-VL as the foundation of our model are as follows: First, Qwen2-VL represents a state-of-the-art open-weight vision-language model that demonstrates outstanding performance in various understanding and detection tasks. Its ability to comprehend detailed features and semantic structures proves particularly beneficial for distortion detection, which necessitates both fine-grained analysis of visual cues and an understanding of semantic relationships (e.g., recognizing the structural integrity of a human hand). Second, the encoder of Qwen2-VL is designed to accept various resolution images, including high-resolution images, which is crucial for capturing subtle distortions in human body parts. High-resolution inputs enable the model to detect minute anomalies that might be missed at lower resolutions, facilitating more accurate and detailed predictions.

By leveraging these advantages, our adapted model effectively detects and predicts distortions in human body parts with improved precision.

\begin{figure*}[h]
\centering
\includegraphics[width=0.85\textwidth]{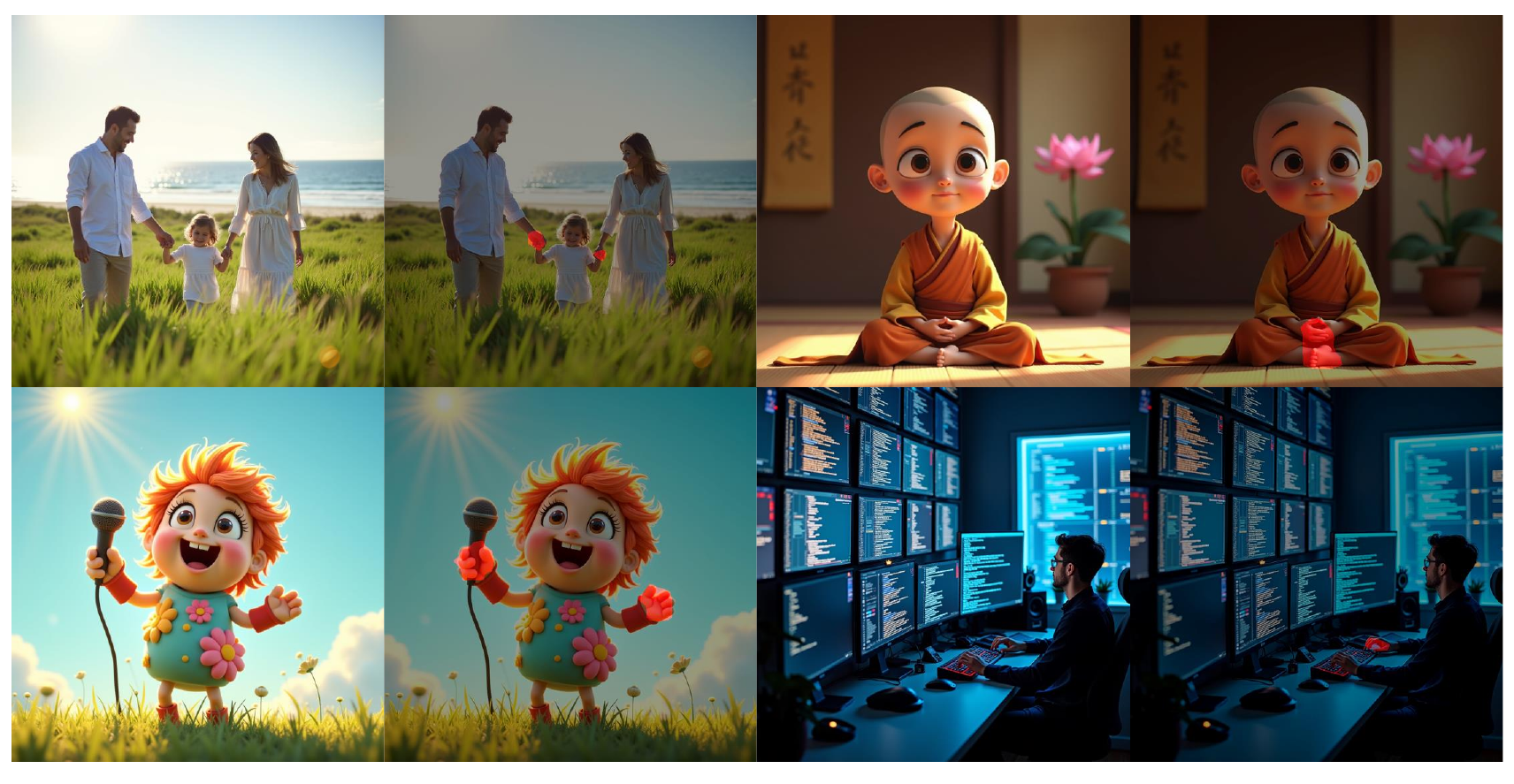}
\caption{\textbf{Examples of test images along with predicted distortion mask.} Our ViT-HD predicts the distorted body parts in these images, where the left image in each pair represents the original, and the right image features red masks indicating the predicted distorted region.
}
\label{fig:prediction_examples}
\end{figure*}

\subsection{Experiments}
\subsubsection{Experimental Details}
\hspace{0.5cm}All models are trained on the 4,000 \emph{Distortion-5K} training samples, and the hyperparameters are tuned using the model performance on the 300 validation samples, and tested on 400 test samples. More details can be found in Appendix \ref{appendix:b}.

We conduct extensive experiments to evaluate the performance of our ViT-HD on the task of distortion detection. To prevent the model from converging to a local optimum where it predominantly predicts easily distorted parts of the human body (such as hands and feet), we employ a two-stage training strategy:

\begin{enumerate}
\item \textbf{Patch-Level Mask Training:} Initially, we train the model using patch-level mask labels. This approach encourages the model to assess distortion from a macro perspective, considering larger regions of the image rather than focusing on localized details. This stage helps the model develop a holistic understanding of potential distortions across various body parts.
\item \textbf{Pixel-Level Mask Fine-Tuning:} Following the initial training, we fine-tune the model using pixel-level mask labels. This step refines the model's ability to localize distortions with greater precision by providing detailed, pixel-wise annotations of distorted regions.
\end{enumerate}

This progressive training strategy allows the model to balance both global and local information, enhancing its overall detection capabilities.

\paragraph{Evaluation Metrics}
Our annotated dataset includes a mask for each image to indicate distorted regions, and we evaluate the models using several standard metrics. Pixel-level precision, recall, and F1 score measure the model's accuracy in predicting distorted pixels, reflecting its ability to correctly identify distorted regions. Additionally, Intersection over Union (IoU) quantifies the overlap between the predicted mask and the ground truth mask, indicating localization accuracy. The Dice coefficient, similar to IoU but more sensitive to small objects, offers an alternative perspective on segmentation performance. Beyond pixel-level metrics, we also assess the models at a macro level to evaluate their ability to detect whether an image contains any distortion. For models capable of predicting masks, an distorted human image is considered correctly detected if the predicted distorted pixels exist in the image.
\paragraph{Baselines}
To benchmark the performance of our model, we compare it against several baseline models:

\begin{itemize}
\item \textbf{Segmentation Models:} We train the standard mask prediction components of U-Net~\cite{u-net} and DeeplabV3~\cite{deeplabv3} to predict the image mask.
\item \textbf{ViT Models:} We fine-tune the Vision Transformer (ViT) ~\cite{alexey2020image} components of CLIP~\cite{clip} and DINOv2~\cite{dinov2}, each equipped with the same MLP head designed to predict distortion masks. These models serve as strong baselines given their proven effectiveness in image representation learning.
\item \textbf{VLMs:} As Visual Language Models (VLMs) cannot predict image masks, we cannot assess their ability through pixel-level metrics. For the advanced proprietary model GPT-4o~\cite{openai2023gpt4,achiam2023gpt}, we conduct tests to distinguish between distorted and normal human images, obtaining image-level metrics. For Qwen2-VL-7B-Instruct~\cite{qwen2-vl}, we also conduct tests to locate distorted regions using rectangular boxes. When testing area-level metrics, we convert each non-adjacent mask in an image into a box label for evaluation.
\end{itemize}

\begin{figure*}[h]
\centering
\begin{subfigure}[b]{0.48\textwidth}
\centering
\includegraphics[width=\linewidth]{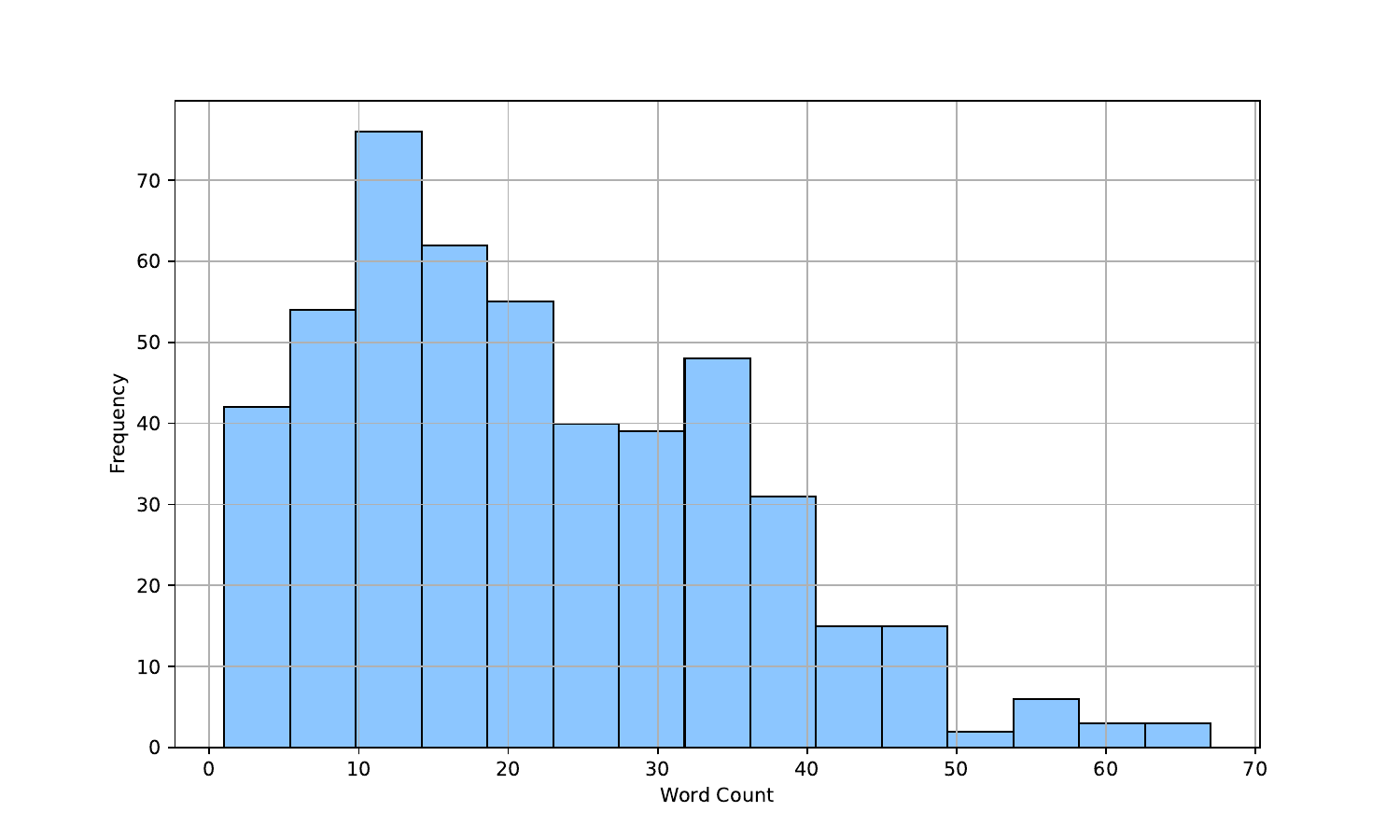}
\caption{The frequency of the word count of the prompts in \emph{Human Distortion Benchmark}.}
\label{fig:benchmark_word_count}
\end{subfigure}
\begin{subfigure}[b]{0.48\textwidth}
\centering
\includegraphics[width=\linewidth]{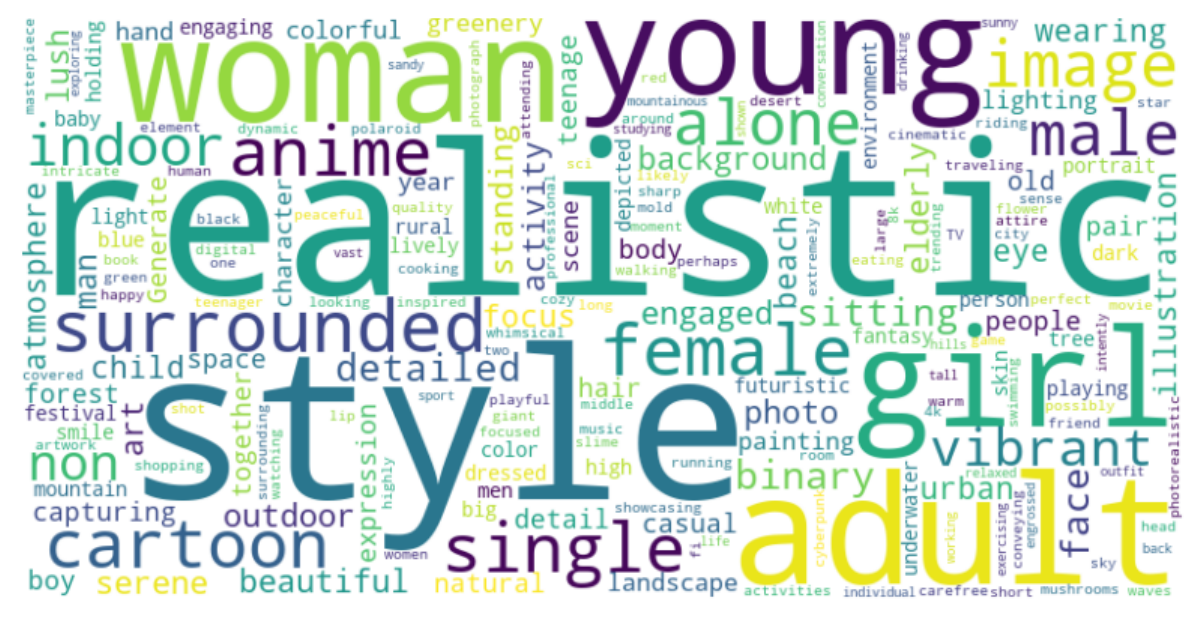}
\caption{All words cloud of the prompts in \emph{Human Distortion Benchmark}.}
\label{fig:wordcloud}
\end{subfigure}
\caption{\textbf{Analysis of our \emph{Human Distortion Benchmark}.} Left: Distribution of word counts. Right: Words cloud.}
\label{fig:benchmark}
\end{figure*}

\subsubsection{Prediction Results on Distortion-5K Test Set}
\paragraph{Quantitative Analysis}
The experimental results of our model's prediction on the distortion mask using the \emph{Distortion-5K} test set are presented in Table \ref{tab:ViT-HD}. The table provides a comprehensive comparison of various models across multiple evaluation metrics, including pixel-level, area-level, and image-level metrics.

Our proposed model, ViT-HD, demonstrates superior performance across all metrics compared to other models. This significant improvement underscores the effectiveness of our approach in addressing the distortion detection problem. Notably, the two-stage training strategy, which involves an initial patch-level training phase followed by pixel-level fine-tuning, proves to be more effective than direct pixel-level training. This suggests that patch-level training helps the model avoid overfitting to local details early in the training process, thereby enabling it to better capture the macroscopic features of distorted regions.

In contrast, GPT-4o, despite its advanced capabilities in other domains, performs poorly in this task. We construct specific prompts to guide GPT-4o in identifying distorted regions within images (see Appendix \ref{appendix:b} for details). However, the model often fails to accurately detect or describe these regions. This limitation may be attributed to inherent hallucinations in its knowledge base and insufficient visual reasoning capabilities ~\cite{bai2024hallucination,sun2025hallucinations,tong2024eyes}. For instance, when presented with an image of a hand clearly with six fingers, GPT-4o incorrectly identifies it as having five fingers, demonstrating its inability to accurately interpret visual anomalies. For Qwen2-VL-7B-Instruct, although recall is high, it classifies almost every image as distorted, which lacks practical value. Even after fine-tuning for the localization task, its area-level metric performance is poor, and while the image-level metric accuracy improves, recall drops sharply. We believe that distortion recognition is a purely visual task, and the language model of Qwen2-VL-7B-Instruct does not aid in this task.

\paragraph{Qualitative Examples}
We present several example predictions from our model for predicting distortion masks. Figure \ref{fig:prediction_examples} shows example images along with predicted distortion mask. Additional examples are available in the supplementary material. These qualitative results illustrate the model's capability to accurately detect and localize distortions in human body parts.

\begin{figure}
    \centering
    \includegraphics[width=0.48\textwidth]{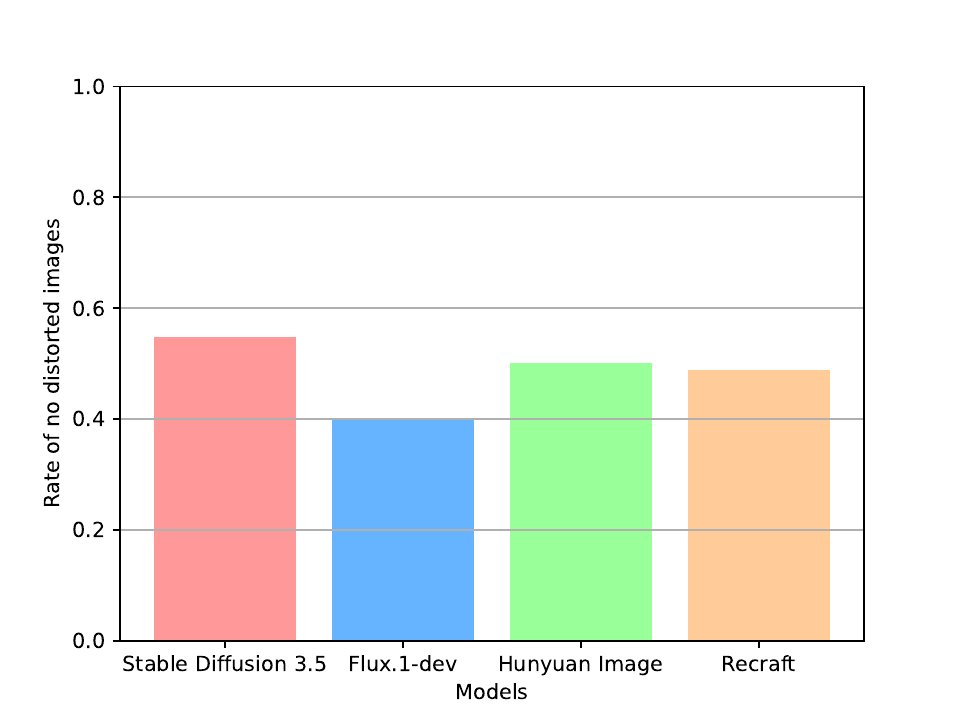}
    \caption{Rate of undistorted AI-generated images on \emph{Human Distortion Benchmark}.}
    \label{fig:benchmark_result1}
\end{figure}

\begin{figure}
    \centering
    \includegraphics[width=0.48\textwidth]{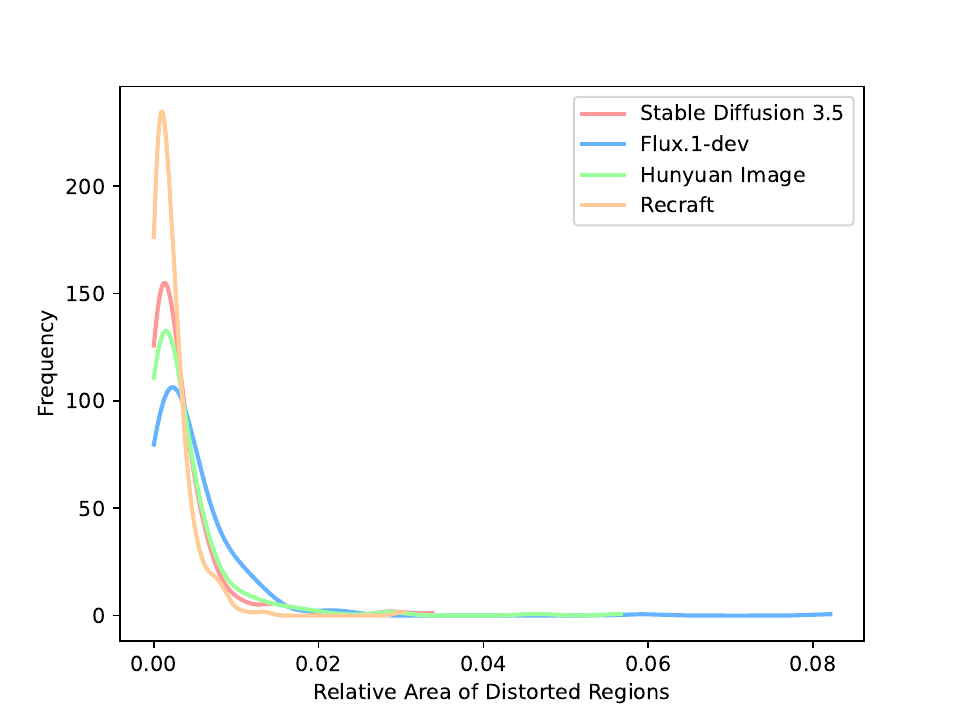}
    \caption{\textbf{Comparison of four state-of-art T2I models on \emph{Human Distortion Benchmark}.} Left: Rate of undistorted images. Right: The distribution of predicted distorted areas.}
    \label{fig:benchmark_result2}
\end{figure}

\section{Evaluating Popular Text-to-Image Models}
This section presents a comprehensive evaluation of popular text-to-image (T2I) models using a newly constructed \emph{Human Distortion Benchmark}. The benchmark comprises 500 human-related T2I prompts, designed to assess the performance of T2I models in generating distortion-free human images. Four widely-used T2I models are evaluated using the proposed Distortion Predicting Model, which provides both quantitative and qualitative insights into the models' capabilities.

\subsection{Benchmark Construction}
The construction of the \emph{Human Distortion Benchmark} relies on a diverse set of prompts, categorized into two main sources: automatically generated prompts and prompts collected from real-world usage. Each category contains 250 prompts, ensuring a balanced representation of synthetic and real-world data.
\paragraph{Automatically Generated Prompts}
To generate synthetic prompts, Llama-3.1-8B ~\cite{llama,llama-3.1} is employed to list meta-attributes commonly associated with human-centric T2I prompts. These attributes include the number of humans, age, gender, artistic style, human activity and so on. Random combinations of these attributes are sampled, then Llama-3.1-8B generates corresponding human-related T2I prompts.

However, Llama-3.1-8B is not entirely perfect for this task, as the generated prompts occasionally exhibit inconsistencies between the meta-attributes. To address this issue, a self-check mechanism is introduced during the construction process. Specifically, Llama-3.1-8B evaluates the semantic similarity between the generated prompts and their corresponding meta-attributes. Prompts that fail to meet a predefined similarity threshold are refined, ensuring the quality and consistency of the benchmark.

\paragraph{Prompts from Real World}
To ensure that the benchmark aligns with real-world user needs, 250 human-related prompts are sampled from the Pick-a-Pic dataset~\cite{kirstain2023pick}, which contains prompts submitted by real users. These prompts are integrated into the benchmark, providing a realistic set of inputs for evaluating T2I models.

\subsection{Benchmark Candidates}
Four popular text-to-image (T2I) models, including both open-source and proprietary options, are evaluated using the \emph{Human Distortion Benchmark}. The models assessed are \textbf{Stable Diffusion 3.5}~\cite{stable-diffusion-3.5-large, rombach2022high}, \textbf{Flux.1-dev}~\cite{flux2024, flux.1-dev}, \textbf{Recraft v3}~\cite{recraft}, and \textbf{Hunyuan Image}~\cite{yuanbao,li2024hunyuan}, representing state-of-the-art performance.

\subsection{Evaluation Results}
To ensure a fair comparison, all models are evaluated under identical conditions. Each model generates images at a resolution of 1024x1024, using the same set of prompts from the benchmark. 

All generated images are evaluated using our ViT-HD metric. We randomly sample 50 images from each model and have an additional expert review the predicted results to examine the effectiveness of our metric. The expert finds that nearly 80\% of these samples are correctly predicted, indicating that our metric possesses a certain degree of generalization, which can be used for further evaluation.

As shown in Figure \ref{fig:benchmark_result1}, the performance of all models is suboptimal, with only Stable Diffusion 3.5 generating images with a non-distortion rate exceeding 0.5, while more than half of the images generated by the other models are distorted. Additionally, we perform a statistical analysis of the relative areas (as a percentage of total image area) of the predicted distorted regions on the \emph{Human Distortion Benchmark}, with the detailed distribution presented in Figure \ref{fig:benchmark_result2}. It can be observed that the distorted areas in the generated images exhibit a long-tail distribution, with most distortions being relatively small in area. Among the models, Hunyuan Image performs better, with fewer images having small distortions and also fewer images with large distortions.

\section{Conclusion}
In this work, we present Distortion-5K, the first dataset covering diverse human distortion AI-generated images, and propose ViT-HD, a Vision Transformer-based model that significantly outperforms various methods in localizing these human disortions. Our experiments demonstrate that current T2I models struggle with anatomical accuracy, as evidenced by the Human Distortion Benchmark, where half of generated images exhibit distortions. While ViT-HD excels in identifying subtle distortion, challenges remain in handling ambiguous distortion categorizations and generalizing to more complex situations. Future work will explore the use of ViT-HD to improve the performance of human figure generation, such as filtering the training dataset, helping refine the generated image. By bridging the gap between generative capabilities and anatomical precision, this research provides a foundation for developing more reliable T2I models, ultimately enhancing their utility in fields requiring human-centric visual content.

{\small
\bibliographystyle{ieee}
\bibliography{egbib}

\begin{thebibliography}{10}\itemsep=-1pt

\bibitem{flux.1-dev}
Flux.1-dev.
\newblock Available at: \url{https://huggingface.co/black-forest-labs/FLUX.1-dev}.

\bibitem{llama-3.1}
llama-3.1.
\newblock \url{https://ai.meta.com/blog/meta-llama-3-1/}.

\bibitem{recraft}
Recraft.
\newblock Available at: \url{https://www.recraft.ai/}.

\bibitem{stable-diffusion-3.5-large}
stable-diffusion-3.5-large.
\newblock Available at: \url{https://huggingface.co/stabilityai/stable-diffusion-3.5-large}.

\bibitem{achiam2023gpt}
J.~Achiam, S.~Adler, S.~Agarwal, L.~Ahmad, I.~Akkaya, F.~L. Aleman, D.~Almeida, J.~Altenschmidt, S.~Altman, S.~Anadkat, et~al.
\newblock Gpt-4 technical report.
\newblock {\em arXiv preprint arXiv:2303.08774}, 2023.

\bibitem{alexey2020image}
D.~Alexey.
\newblock An image is worth 16x16 words: Transformers for image recognition at scale.
\newblock {\em arXiv preprint arXiv: 2010.11929}, 2020.

\bibitem{bai2024hallucination}
Z.~Bai, P.~Wang, T.~Xiao, T.~He, Z.~Han, Z.~Zhang, and M.~Z. Shou.
\newblock Hallucination of multimodal large language models: A survey.
\newblock {\em arXiv preprint arXiv:2404.18930}, 2024.

\bibitem{barratt2018note}
S.~Barratt and R.~Sharma.
\newblock A note on the inception score.
\newblock {\em arXiv preprint arXiv:1801.01973}, 2018.

\bibitem{chen2023pixart}
J.~Chen, J.~Yu, C.~Ge, L.~Yao, E.~Xie, Y.~Wu, Z.~Wang, J.~Kwok, P.~Luo, H.~Lu, et~al.
\newblock Pixart: Fast training of diffusion transformer for photorealistic text-to-image synthesis.
\newblock {\em arXiv preprint arXiv:2310.00426}, 2023.

\bibitem{deeplabv3}
L.-C. Chen.
\newblock Rethinking atrous convolution for semantic image segmentation.
\newblock {\em arXiv preprint arXiv:1706.05587}, 2017.

\bibitem{chen2024overview}
M.~Chen, S.~Mei, J.~Fan, and M.~Wang.
\newblock An overview of diffusion models: Applications, guided generation, statistical rates and optimization.
\newblock {\em arXiv preprint arXiv:2404.07771}, 2024.

\bibitem{chen2023exploring}
Z.~Chen, W.~Sun, H.~Wu, Z.~Zhang, J.~Jia, Z.~Ji, F.~Sun, S.~Jui, X.~Min, G.~Zhai, et~al.
\newblock Exploring the naturalness of ai-generated images.
\newblock {\em arXiv preprint arXiv:2312.05476}, 2023.

\bibitem{dhariwal2021diffusion}
P.~Dhariwal and A.~Nichol.
\newblock Diffusion models beat gans on image synthesis.
\newblock {\em Advances in neural information processing systems}, 34:8780--8794, 2021.

\bibitem{ding2022cogview2}
M.~Ding, W.~Zheng, W.~Hong, and J.~Tang.
\newblock Cogview2: Faster and better text-to-image generation via hierarchical transformers.
\newblock {\em Advances in Neural Information Processing Systems}, 35:16890--16902, 2022.

\bibitem{yuanbao}
J.~Doe.
\newblock Recraft.
\newblock Available at: \url{https://yuanbao.tencent.com/}.

\bibitem{llama}
A.~Dubey, A.~Jauhri, A.~Pandey, A.~Kadian, A.~Al-Dahle, A.~Letman, A.~Mathur, A.~Schelten, A.~Yang, A.~Fan, et~al.
\newblock The llama 3 herd of models.
\newblock {\em arXiv preprint arXiv:2407.21783}, 2024.

\bibitem{gandikota2025concept}
R.~Gandikota, J.~Materzy{\'n}ska, T.~Zhou, A.~Torralba, and D.~Bau.
\newblock Concept sliders: Lora adaptors for precise control in diffusion models.
\newblock In {\em European Conference on Computer Vision}, pages 172--188. Springer, 2025.

\bibitem{goodfellow2014generative}
I.~Goodfellow, J.~Pouget-Abadie, M.~Mirza, B.~Xu, D.~Warde-Farley, S.~Ozair, A.~Courville, and Y.~Bengio.
\newblock Generative adversarial nets.
\newblock {\em Advances in neural information processing systems}, 27, 2014.

\bibitem{herz2019understanding}
M.~Herz and P.~A. Rauschnabel.
\newblock Understanding the diffusion of virtual reality glasses: The role of media, fashion and technology.
\newblock {\em Technological Forecasting and Social Change}, 138:228--242, 2019.

\bibitem{heusel2017gans}
M.~Heusel, H.~Ramsauer, T.~Unterthiner, B.~Nessler, and S.~Hochreiter.
\newblock Gans trained by a two time-scale update rule converge to a local nash equilibrium.
\newblock {\em Advances in neural information processing systems}, 30, 2017.

\bibitem{ho2020denoising}
J.~Ho, A.~Jain, and P.~Abbeel.
\newblock Denoising diffusion probabilistic models.
\newblock {\em Advances in neural information processing systems}, 33:6840--6851, 2020.

\bibitem{kingma2013auto}
D.~P. Kingma.
\newblock Auto-encoding variational bayes.
\newblock {\em arXiv preprint arXiv:1312.6114}, 2013.

\bibitem{kirstain2023pick}
Y.~Kirstain, A.~Polyak, U.~Singer, S.~Matiana, J.~Penna, and O.~Levy.
\newblock Pick-a-pic: An open dataset of user preferences for text-to-image generation.
\newblock {\em Advances in Neural Information Processing Systems}, 36:36652--36663, 2023.

\bibitem{flux2024}
B.~F. Labs.
\newblock Flux.
\newblock \url{https://github.com/black-forest-labs/flux}, 2024.

\bibitem{li2024aigiqa}
C.~Li, T.~Kou, Y.~Gao, Y.~Cao, W.~Sun, Z.~Zhang, Y.~Zhou, Z.~Zhang, W.~Zhang, H.~Wu, et~al.
\newblock Aigiqa-20k: A large database for ai-generated image quality assessment.
\newblock {\em arXiv preprint arXiv:2404.03407}, 2(3):5, 2024.

\bibitem{li2024hunyuan}
Z.~Li, J.~Zhang, Q.~Lin, J.~Xiong, Y.~Long, X.~Deng, Y.~Zhang, X.~Liu, M.~Huang, Z.~Xiao, et~al.
\newblock Hunyuan-dit: A powerful multi-resolution diffusion transformer with fine-grained chinese understanding.
\newblock {\em arXiv preprint arXiv:2405.08748}, 2024.

\bibitem{liang2024rich}
Y.~Liang, J.~He, G.~Li, P.~Li, A.~Klimovskiy, N.~Carolan, J.~Sun, J.~Pont-Tuset, S.~Young, F.~Yang, et~al.
\newblock Rich human feedback for text-to-image generation.
\newblock In {\em Proceedings of the IEEE/CVF Conference on Computer Vision and Pattern Recognition}, pages 19401--19411, 2024.

\bibitem{lu2024handrefiner}
W.~Lu, Y.~Xu, J.~Zhang, C.~Wang, and D.~Tao.
\newblock Handrefiner: Refining malformed hands in generated images by diffusion-based conditional inpainting.
\newblock In {\em Proceedings of the 32nd ACM International Conference on Multimedia}, pages 7085--7093, 2024.

\bibitem{openai2023gpt4}
OpenAI.
\newblock Gpt-4, 2023.
\newblock Accessed: 2023-10-01.

\bibitem{dinov2}
M.~Oquab, T.~Darcet, T.~Moutakanni, H.~Vo, M.~Szafraniec, V.~Khalidov, P.~Fernandez, D.~Haziza, F.~Massa, A.~El-Nouby, et~al.
\newblock Dinov2: Learning robust visual features without supervision.
\newblock {\em arXiv preprint arXiv:2304.07193}, 2023.

\bibitem{clip}
A.~Radford, J.~W. Kim, C.~Hallacy, A.~Ramesh, G.~Goh, S.~Agarwal, G.~Sastry, A.~Askell, P.~Mishkin, J.~Clark, et~al.
\newblock Learning transferable visual models from natural language supervision.
\newblock In {\em International conference on machine learning}, pages 8748--8763. PMLR, 2021.

\bibitem{ramesh2021zero}
A.~Ramesh, M.~Pavlov, G.~Goh, S.~Gray, C.~Voss, A.~Radford, M.~Chen, and I.~Sutskever.
\newblock Zero-shot text-to-image generation.
\newblock In {\em International conference on machine learning}, pages 8821--8831. Pmlr, 2021.

\bibitem{rombach2022high}
R.~Rombach, A.~Blattmann, D.~Lorenz, P.~Esser, and B.~Ommer.
\newblock High-resolution image synthesis with latent diffusion models.
\newblock In {\em Proceedings of the IEEE/CVF conference on computer vision and pattern recognition}, pages 10684--10695, 2022.

\bibitem{u-net}
O.~Ronneberger, P.~Fischer, and T.~Brox.
\newblock U-net: Convolutional networks for biomedical image segmentation.
\newblock In {\em Medical image computing and computer-assisted intervention--MICCAI 2015: 18th international conference, Munich, Germany, October 5-9, 2015, proceedings, part III 18}, pages 234--241. Springer, 2015.

\bibitem{saharia2022photorealistic}
C.~Saharia, W.~Chan, S.~Saxena, L.~Li, J.~Whang, E.~L. Denton, K.~Ghasemipour, R.~Gontijo~Lopes, B.~Karagol~Ayan, T.~Salimans, et~al.
\newblock Photorealistic text-to-image diffusion models with deep language understanding.
\newblock {\em Advances in neural information processing systems}, 35:36479--36494, 2022.

\bibitem{sun2025hallucinations}
S.~Sun, Z.~Lin, and X.~Wu.
\newblock Hallucinations of large multimodal models: Problem and countermeasures.
\newblock {\em Information Fusion}, page 102970, 2025.

\bibitem{tong2024eyes}
S.~Tong, Z.~Liu, Y.~Zhai, Y.~Ma, Y.~LeCun, and S.~Xie.
\newblock Eyes wide shut? exploring the visual shortcomings of multimodal llms.
\newblock In {\em Proceedings of the IEEE/CVF Conference on Computer Vision and Pattern Recognition}, pages 9568--9578, 2024.

\bibitem{van2017neural}
A.~Van Den~Oord, O.~Vinyals, et~al.
\newblock Neural discrete representation learning.
\newblock {\em Advances in neural information processing systems}, 30, 2017.

\bibitem{wang2024diffusion}
B.~Wang, Q.~Chen, and Z.~Wang.
\newblock Diffusion-based visual art creation: A survey and new perspectives.
\newblock {\em arXiv preprint arXiv:2408.12128}, 2024.

\bibitem{wang2024rhands}
C.~Wang, P.~Liu, M.~Zhou, M.~Zeng, X.~Li, T.~Ge, et~al.
\newblock Rhands: Refining malformed hands for generated images with decoupled structure and style guidance.
\newblock {\em arXiv preprint arXiv:2404.13984}, 2024.

\bibitem{qwen2-vl}
P.~Wang, S.~Bai, S.~Tan, S.~Wang, Z.~Fan, J.~Bai, K.~Chen, X.~Liu, J.~Wang, W.~Ge, et~al.
\newblock Qwen2-vl: Enhancing vision-language model's perception of the world at any resolution.
\newblock {\em arXiv preprint arXiv:2409.12191}, 2024.

\bibitem{weng2023diffusion}
Z.~Weng, L.~Bravo-S{\'a}nchez, and S.~Yeung.
\newblock Diffusion-hpc: Generating synthetic images with realistic humans.
\newblock {\em arXiv preprint arXiv:2303.09541}, 2, 2023.

\bibitem{wu2023humanv2}
X.~Wu, Y.~Hao, K.~Sun, Y.~Chen, F.~Zhu, R.~Zhao, and H.~Li.
\newblock Human preference score v2: A solid benchmark for evaluating human preferences of text-to-image synthesis.
\newblock {\em arXiv preprint arXiv:2306.09341}, 2023.

\bibitem{wu2023human}
X.~Wu, K.~Sun, F.~Zhu, R.~Zhao, and H.~Li.
\newblock Human preference score: Better aligning text-to-image models with human preference.
\newblock In {\em Proceedings of the IEEE/CVF International Conference on Computer Vision}, pages 2096--2105, 2023.

\bibitem{xu2024imagereward}
J.~Xu, X.~Liu, Y.~Wu, Y.~Tong, Q.~Li, M.~Ding, J.~Tang, and Y.~Dong.
\newblock Imagereward: Learning and evaluating human preferences for text-to-image generation.
\newblock {\em Advances in Neural Information Processing Systems}, 36, 2024.

\bibitem{yang2023diffusion}
L.~Yang, Z.~Zhang, Y.~Song, S.~Hong, R.~Xu, Y.~Zhao, W.~Zhang, B.~Cui, and M.-H. Yang.
\newblock Diffusion models: A comprehensive survey of methods and applications.
\newblock {\em ACM Computing Surveys}, 56(4):1--39, 2023.

\bibitem{ye2023affordance}
Y.~Ye, X.~Li, A.~Gupta, S.~De~Mello, S.~Birchfield, J.~Song, S.~Tulsiani, and S.~Liu.
\newblock Affordance diffusion: Synthesizing hand-object interactions.
\newblock In {\em Proceedings of the IEEE/CVF Conference on Computer Vision and Pattern Recognition}, pages 22479--22489, 2023.

\bibitem{zhang2024learning}
S.~Zhang, B.~Wang, J.~Wu, Y.~Li, T.~Gao, D.~Zhang, and Z.~Wang.
\newblock Learning multi-dimensional human preference for text-to-image generation.
\newblock In {\em Proceedings of the IEEE/CVF Conference on Computer Vision and Pattern Recognition}, pages 8018--8027, 2024.

\end{thebibliography}
}

\newpage
\appendix
\section{Data Collection Details}
\label{appendix:a}
\subsection{Definition of Human Distortion}
Human body distortion refers to abnormalities in the anatomical structure of the human body that deviate from normal physiological standards. These distortions manifest in various forms, including proliferation, absence, deformation, and fusion. Each category is defined and exemplified below to provide a comprehensive understanding of the phenomena under study.

\textbf{Proliferation:} This refers to the abnormal increase in the number of body parts beyond the normal physiological count or the appearance of body parts in atypical locations. Examples include extra fingers (e.g., polydactyly), additional arms, or duplicated toes in the extremities; multiple eyes (e.g., triophthalmia), extra ears, or additional facial features such as noses or mouths in the head; and duplicated chest or abdominal structures.

\textbf{Absence:} This refers to the abnormal lack or reduction in the number of body parts, excluding cases caused by perspective or angle-related issues. Examples include missing fingers (e.g., oligodactyly), absent arms, or a reduced number of toes in the extremities; lack of ears (e.g., anotia), missing eyes (e.g., anophthalmia), or absent facial features such as noses or mouths in the head; and the absence of chest or abdominal structures.

\textbf{Deformation:} This refers to the abnormal distortion or warping of body parts into shapes, sizes, or proportions that deviate from natural physiological standards. Examples include disproportionate hand or foot sizes (e.g., macrodactyly), elongated or shortened fingers or toes, or irregular limb proportions in the extremities; misaligned or asymmetrical facial features, such as uneven eyes, distorted noses, or irregularly shaped mouths in the head; and abnormal chest or abdominal contours.

\textbf{Fusion:} This refers to the unnatural merging or overlapping of body parts, either within a single individual or between multiple individuals or objects. Examples include fused fingers (e.g., syndactyly), overlapping limbs, or intertwined legs in the extremities; merged facial features, such as a hand fused to the face or overlapping facial structures in the head; and unnatural adhesion of chest or abdominal regions, or the fusion of the torso with external objects in the torso.

\subsection{Annotation Details}
We conduct multiple rounds of training for the annotators, utilizing the aforementioned deformity criteria along with representative examples of each type of distortion. Initially, a subset of samples is annotated as a trial. These annotations are then evaluated by experts, who provide feedback on the problematic cases. Following this expert assessment, the annotators undergo additional training before continuing with the annotation process. Throughout this iterative process, we obtain several important guidelines:
\begin{enumerate}
\item If the human body can assume the position depicted in the image, even if it appears highly unnatural, it should not be considered a deformity.
\item When encountering ambiguous body parts, annotators can attempt to replicate the pose themselves (e.g., the position of fingers). If the pose can be naturally assumed, it should not be classified as a deformity.
\item Regions that are too blurred to make a clear judgment should not be annotated as deformities.
\item The tolerance for deformities varies with different image styles. For instance, in realistic images, the threshold for identifying deformities is lower, such as considering the absence of visible finger joints as a deformity. However, in anime-style images, where hand joints are sometimes not clearly depicted, such details do not need to be annotated. It is important to distinguish between these styles.
\end{enumerate}

\begin{figure*}[!ht]
\centering
\includegraphics[width=0.98\textwidth]{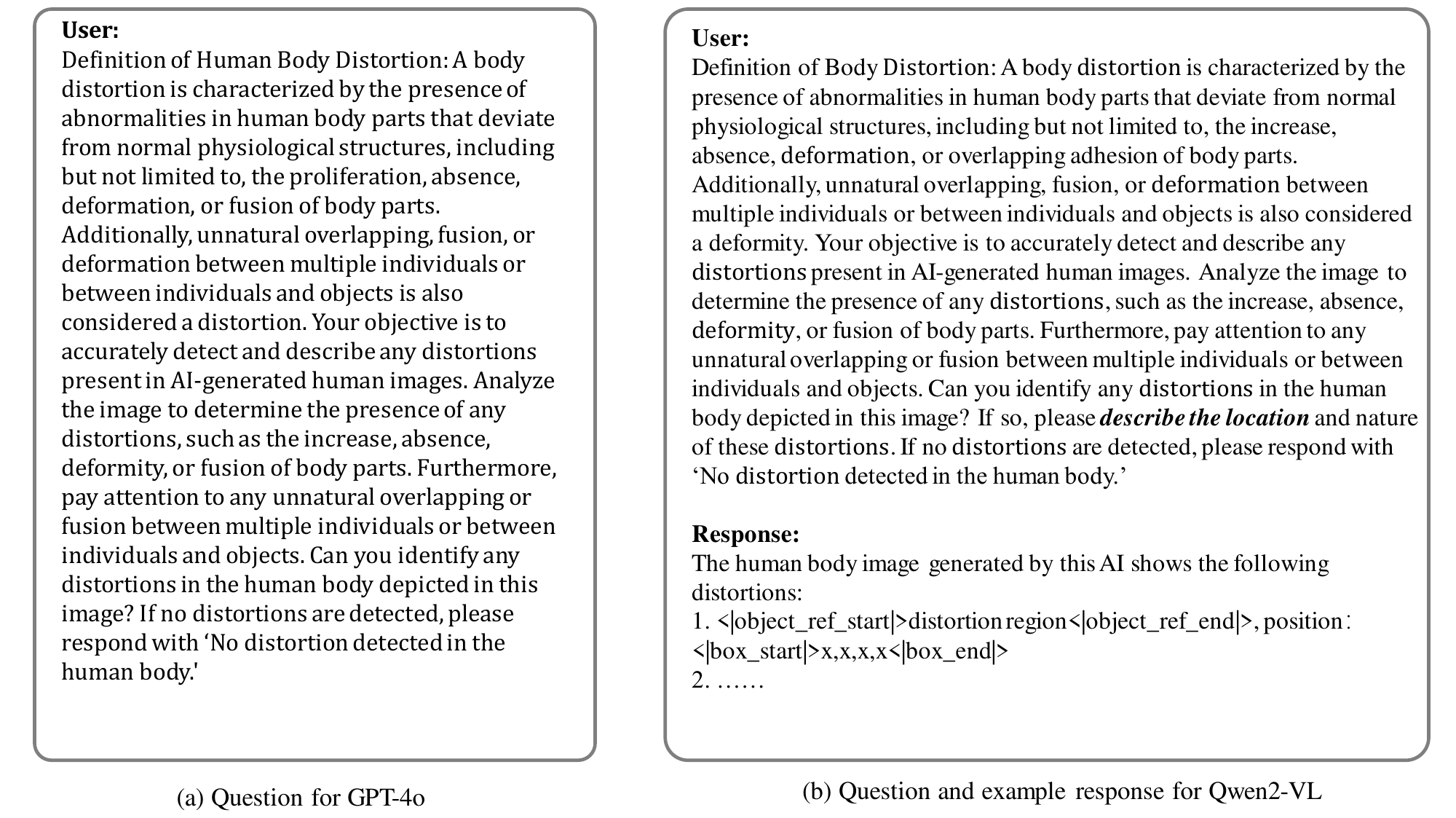}
\caption{\textbf{Prompts for GPT-4o and Qwen2-VL}. Left: Question posed for GPT-4o. Right: Question posed for Qwen2-VL and the corresponding answer to fine-tune it.
}
\label{fig:prompt}
\end{figure*}

\section{Experimental Details}
\label{appendix:b}
\subsection{Details of Segmentation and ViT Models}
All models are trained on the datasets with a consistent batch size of 32 for 8 epochs. The optimal learning rate is determined through a grid search over the range [1e-3, 5e-3, 1e-4, 5e-4, 1e-5, 5e-5, 1e-6], with performance evaluated on the validation set. During training, a weight decay of 0.01 is applied to regularize the model parameters and prevent overfitting. The learning rate schedule follows a linear decay strategy, with a warmup phase accounting for 10\% of the total training steps to stabilize the initial training process.

The U-Net architecture, a convolutional neural network specifically designed for image segmentation tasks, comprises 10 convolutional layers and 4 deconvolutional layers organized in a symmetric encoder-decoder structure. For semantic segmentation, we employ the DeepLabV3 model, which utilizes a ResNet-50 backbone for feature extraction and incorporates an atrous spatial pyramid pooling (ASPP) module to effectively capture multi-scale contextual information. This architecture consists of the convolutional layers from the ResNet-50 backbone followed by a classifier that generates the final segmentation map.

The CLIP implementation used in this study is the 'clip-vit-large-patch14-336' variant, which is augmented with a 2-layer multilayer perceptron (MLP) and bilinear interpolation. Due to computational memory constraints, both input images and their corresponding masks are resized to 560 pixels. Similarly, the DINOv2 model employed is the 'dinov2-giant' variant, combined with an identical pixel-level MLP architecture as used in the CLIP implementation. The ViT-HD model is initialized using weights from the 'Qwen2-VL-Instruct' pretrained model.

The training procedure is conducted in two distinct stages. During patch-level training, patch size is 14x14; image patches are classified as distorted if more than 50\% of their area is identified as distorted. The model is trained for 3 epochs. In the subsequent pixel-level training, each pixel is individually labeled, and training proceeds until convergence criteria are met. This hierarchical approach ensures both global context understanding and precise pixel-level accuracy in the final segmentation output.
\subsection{Details of VLMs}
Since GPT-4o is incapable of directly predicting the location of distorted parts, we instead query whether there are any distortions in an entire picture and evaluate its performance in distorted body recognition at the whole-picture level. The prompts we employ are presented in Figure \ref{fig:prompt}, as detailed below.

Qwen2-VL-7B-Instruct is trained with localization tasks and thus can identify the bounding boxes of objects. Consequently, we not only ask Qwen2-VL to assess the distortion situation of the entire picture but also require it to indicate the position of the distorted area. During the fine-tuning process, the coordinates of the rectangular bounding boxes that precisely enclose each non-adjacent mask are used in the responses. The questions and answering prompts we utilize are also shown in Figure \ref{fig:prompt}.

\end{document}